\documentclass{article}
\usepackage{ijcai26}

\usepackage{times}
\usepackage{soul}
\usepackage{url}
\usepackage[hidelinks]{hyperref}
\usepackage[utf8]{inputenc}
\usepackage[small]{caption}
\usepackage{graphicx}
\usepackage{amsmath}
\usepackage{amsthm}
\usepackage{booktabs}
\usepackage{algorithm}
\usepackage{algorithmic}
\usepackage[switch]{lineno}

\title{Radiologist-Guided Causal Concept Bottleneck Models for Chest X-Ray Interpretation}

\author{
Amy Rafferty$^1$
\and
Rishi Ramaesh$^2$\and
Ajitha Rajan$^1$\\
\affiliations
$^1$University of Edinburgh, UK\\
$^2$NHS Lothian, UK\\
\emails
s1817812@ed.ac.uk
}

\begin{document}

\maketitle

\begin{abstract}
Concept Bottleneck Models (CBMs) in medical imaging aim to improve model interpretability by predicting intermediate clinical concepts before final diagnoses. However, most existing CBMs treat concepts as discriminative predictors of pathology labels, without explicitly modelling the underlying clinical generative process where diseases produce observable radiographic findings. We propose XpertCausal, a radiologist-guided causal CBM for chest X-ray interpretation which models pathology-to-concept relationships using a probabilistic noisy-OR framework. This generative model is then inverted via Bayesian inference to estimate pathology probabilities from predicted concepts. Radiologist-curated concept-pathology associations are used to constrain model structure to radiologist-defined clinically plausible reasoning pathways. We evaluate XpertCausal on MIMIC-CXR across pathology classification performance, calibration, explanation quality, and alignment with radiologist-defined reasoning pathways. Compared with both a non-causal CBM baseline and a causal ablation with unconstrained learned associations, XpertCausal achieves improved AUROC, calibration, and clinically relevant explanation quality, while learning concept–pathology relationships that more closely align with expert knowledge. These results demonstrate that incorporating clinically motivated causal structure and expert domain knowledge into CBMs can lead to more accurate, interpretable, and clinically aligned models for CXR interpretation.
\end{abstract}

\section{Introduction}

Chest X-rays (CXRs) are one of the most commonly performed medical imaging examinations worldwide, and serve as a first-line diagnostic tool in the diagnosis of thoracic diseases such as pneumonia and lung cancer. Recent advances in deep learning have led to significant improvements in automated CXR interpretation, with convolutional neural networks achieving expert-level performance on a range of pathology classification tasks \cite{ai_surpass,ai_surpass2}. However, despite strong predictive performance, most existing approaches operate as black-box models, providing limited insight into the reasoning processes underlying their predictions. This lack of transparency remains a major barrier to clinical deployment, particularly in high-stakes clinical settings where interpretability and trust are essential, and incorrect predictions can lead to serious patient harm \cite{blackboxbad,blackbox}.

Concept Bottleneck Models (CBMs) have emerged as a promising approach for improving interpretability in deep learning systems \cite{cbms}. Rather than directly predicting labels from images, CBMs first predict a set of human-interpretable concepts before using these concepts to produce final predictions. In the context of CXR interpretation, these concepts are typically observable image features, such as areas of opacity or heart enlargement. This intermediate concept layer provides \textit{explanations} of model predictions in clinically meaningful terms, allowing them to be inspected and validated by end users (i.e., clinicians). 

CBMs have increasingly been explored for medical imaging tasks \cite{medicalcbm,graphcbm}. However, their application is often constrained by the limited availability of large-scale datasets with ground truth concept-level annotations. Obtaining these annotations from radiologists is expensive, time-consuming, and difficult to scale, particularly for complex medical imaging datasets. As a result, many recent approaches instead rely on weakly supervised, automatically generated, or latent concepts \cite{lcbms,posthoccbm,semicbm}, including attention-based concept discovery methods \cite{xcbs} and large language model (LLM)-based semantic representations \cite{labo}. While such approaches can improve interpretability compared with fully black-box models, their learned concepts are often not explicitly clinically grounded, semantically stable, or aligned with meaningful diagnostic reasoning processes \cite{conceptbad,clinicxai}. 

In response to this, XpertXAI \cite{xpertxai} recently demonstrated that clinically grounded concepts can instead be extracted directly from radiology reports using radiologist-defined phrase mappings. By leveraging expert-defined findings rather than learned latent representations, XpertXAI provided a concept bottleneck framework whose intermediate reasoning process aligns more closely with clinically meaningful radiographic observations.

Despite these advantages, most existing CBMs are fundamentally discriminative, where concepts are typically treated as predictive features used to infer pathology labels \cite{cbms,cbm2}. In real-world clinical reasoning, however, the underlying disease generation process proceeds in the opposite direction: pathologies instead \textit{cause} radiographic findings, which clinicians then interpret to infer diagnoses \cite{qmrprob}. Ignoring this generative structure may encourage models to learn spurious associations between concepts and labels, particularly in biased or noisy medical datasets \cite{shortcut,causalhealth}. Another limitation is that many concept-based models treat concepts as independent predictive variables, overlooking the structured dependencies that naturally exist between diseases and their observable manifestations.

Recent work has explored incorporating causal structure into concept-based learning through causal and structured CBMs \cite{countercbm,causalcbm}. Unlike standard CBMs, which typically treat concepts as independent features, these approaches model dependencies between concepts and downstream predictions using directed graphical structures, structural causal models (SCMs), or probabilistic generative frameworks \cite{causalrepr} in order to improve explanation faithfulness, robustness, and alignment between model reasoning and human-understandable concepts \cite{scm,structurecbm,gencbm}. However, many existing approaches either infer these relationships directly from observational data or evaluate primarily on simplified benchmark settings with synthetic or non-clinical concepts. In medical imaging, these learned relationships may instead capture spurious correlations or clinically implausible dependencies rather than real diagnostic reasoning pathways.

Probabilistic graphical models have a long history in computer-assisted medical diagnosis. Early systems such as INTERNIST-1 and Quick Medical Reference (QMR) explored structured representations of disease–finding relationships for diagnostic reasoning \cite{internist,qmr}, while probabilistic alternatives such as QMR-DT modelled diseases and clinical findings using Bayesian networks and noisy-OR relationships \cite{qmrprob}. In these systems, latent diseases generated observable findings, and posterior disease probabilities were inferred using Bayesian reasoning. Although highly interpretable, these approaches predated modern deep learning and therefore lacked the ability to learn visual representations directly from medical images.

In this work, we propose \textbf{XpertCausal}, a radiologist-guided causal CBM for chest X-ray interpretation that combines interpretable concept bottlenecks with probabilistic causal disease modelling. XpertCausal models pathology-to-concept relationships using a noisy-OR generative framework constrained by radiologist-defined concept-pathology associations. Rather than treating concepts solely as predictive features (Concept $\rightarrow$ Pathology), the model explicitly represents how latent diseases generate observable radiographic findings (Pathology $\rightarrow$ Concept). Final pathology predictions are then obtained through Bayesian inference over this generative framework. By incorporating expert-defined structural constraints, XpertCausal encourages clinically grounded reasoning while reducing implausible concept-pathology associations which may be learned from data alone.

We evaluate XpertCausal on the MIMIC-CXR dataset \cite{PhysioNet-mimic,physionet,mimic} across pathology classification performance, calibration, explanation quality, and alignment with radiologist-defined reasoning pathways. We compare against both a non-causal expert-driven CBM baseline (XpertXAI \cite{xpertxai}) and a causal ablation model with unconstrained learned associations. Our results suggest that incorporating both causal structure and expert-guided constraints improves predictive performance, calibration, and the clinical relevance of model explanations.

The main contributions of this work are as follows:

\begin{itemize}
\item We propose \textbf{XpertCausal}, a radiologist-guided causal CBM for chest X-ray interpretation which combines interpretable concept bottlenecks with probabilistic causal disease modelling using a noisy-OR generative framework and Bayesian inference.
\item We incorporate radiologist-defined concept-pathology associations as structural constraints within the causal graph, enabling clinically grounded pathology-to-finding reasoning and reducing implausible learned relationships.
\item We evaluate the impact of causal structure and expert-guided constraints on pathology prediction performance, calibration, explanation quality, and  alignment with radiologist-defined concept-pathology relationships on MIMIC-CXR.
\end{itemize}

\begin{table*}[tb]
    \centering
    \small
    \begin{tabular}{lrrrrrr}
        \toprule
         & \textbf{No Rel. Finding} & \textbf{Susp. Malignancy} & \textbf{Pneumonia} & \textbf{Pleural Effusion} & \textbf{Cardiac Failure} & \textbf{Pneumothorax} \\
        \midrule
        Unremarkable & 2 & - &- &- &- & - \\
        Mass & - & 2 & - & 1 & -& -\\
        Nodular Disease & - & 2 & 1 & - & -& -\\
        Irregular H/M & - & 2 & - & - & 1& -\\
        Pneumonitis & - & - & 2 &-  & -&- \\
        Consolidation & - & - & 2 & - &- & -\\
        Opacities & - & 1 & 1 & - & 1&- \\
        Infection & - & - & 2 & - & -& -\\
        Pleural Disease & - & 1 & 1 & 2 &1 &1 \\
        Enlarged Heart & - & - & - &-  & 2 & -\\
        Abs. Markings & - & - & - & 1 & -&2 \\
        \bottomrule
    \end{tabular}
    \caption{Radiologist-defined concept-pathology association matrix. Each entry indicates the association strength of a concept and a pathology - strong (2), weak (1), or none (-). Abbreviations: Irregular H/M = Irregular Hilum/Mediastinum; Abs. Markings = Absent Lung Markings.}
    \label{tab:expert_matrix}
\end{table*}

\section{Data Setup}\label{sec:data}

This work uses a subset of PhysioNet’s public MIMIC-CXR dataset of CXRs and associated radiology reports \cite{mimic,PhysioNet-mimic,physionet}. The dataset is restricted to PA (posterior-anterior) images only, as this is the standard viewpoint \cite{ap_pa}, and a single image is selected per study in order to prevent data leakage across patients. The original MIMIC-CXR pathology labels are automatically generated using the CheXpert labeller \cite{chexpert}, and have been shown to be noisy and prone to contextual errors \cite{labelnoise,chexpertplusplus,unreliable3}. Following XpertXAI \cite{xpertxai}, we instead derive both concept annotations and pathology labels directly from the associated radiology reports using a rule-based extraction pipeline, using radiologist-curated phrase lists and negation handling in order to map report text to binary vectors of clinical concepts. Before phrase mapping, reports are cleaned by first extracting the \textit{Findings} and \textit{Impression} sections to reduce confusion caused by irrelevant sections (e.g., \textit{History}, \textit{Comparison}). All code, including the original radiologist-curated phrase lists, are publicly available \footnote{https://anonymous.4open.science/r/XpertCausal-4D74/}.

The concept space used in this work builds on XpertXAI’s predefined set of clinically meaningful findings, further refined by a board-certified radiologist. Closely related concepts are grouped together to reduce sparsity and overlap, resulting in a concept space of 11 concepts. Each CXR may be associated with multiple pathologies. We focus on six target labels: \textit{No Relevant Finding}, \textit{Suspicious Malignancy}, \textit{Cardiac Failure}, \textit{Pleural Effusion}, \textit{Pneumothorax}, and \textit{Pneumonia}. The \textit{No Relevant Finding} label is defined as the absence of all other pathologies. To address class imbalance, we apply One-Sided Selection \cite{onesided} to undersample the majority class (\textit{No Relevant Finding}) while preserving the structure of the minority classes. The resulting dataset of 40,083 CXR-report pairs is split into training, validation, and test sets in an 80/10/10 ratio. Each sample is associated with a binary concept vector indicating the presence of the 11 concepts extracted from the report.

\subsection{Concept-Pathology Associations}\label{sec:associations}

In order to model the clinical relationships between concepts and pathologies, we present a radiologist-curated matrix which specifies the level of association between each concept-pathology pair (Table \ref{tab:expert_matrix}). Each association is labelled as Strong (2), Weak (1), or None (-). Although \textit{No Relevant Finding} is not itself a pathology, we treat it as a mutually exclusive diagnostic state associated with the \textit{Unremarkable} concept. This matrix provides a structured representation of radiologist-defined clinical domain knowledge, and is used both to define the causal relationships in XpertCausal, and to assess how closely modelled associations align with expert-defined relationships.

\begin{figure*}[tbh]
    \centering
    \includegraphics[width=\linewidth]{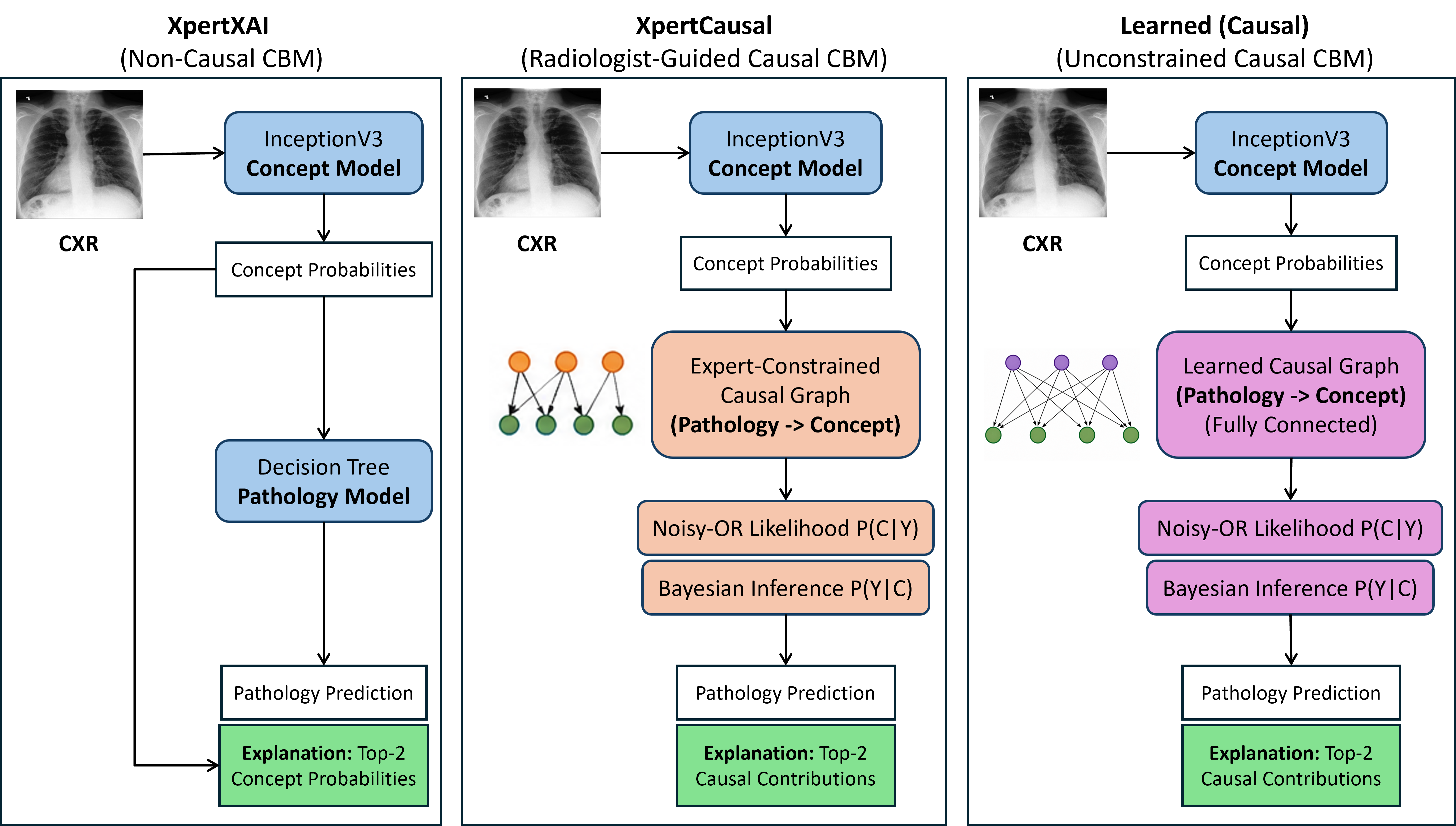}
    \caption{Comparison of model architectures. All models share a common InceptionV3 concept prediction model which maps chest X-rays to clinical concept probabilities. XpertXAI (left) treats concepts as independent features and predicts pathologies using a decision tree. XpertCausal (centre) models the causal relationship between pathologies and concepts using a radiologist-defined graph (pathology $\rightarrow$ concept), and performs prediction via a noisy-OR likelihood combined with Bayesian inference. Causal (Learned) (right) uses the same probabilistic framework but learns all concept–pathology associations without expert constraints, resulting in a fully connected graph. Explanations are derived from the concept space for all models, but differ in how concept importance is determined: predicted probabilities for XpertXAI, and causal contributions for both causal models.}
    \label{fig:overview}
\end{figure*}

\section{Methods}

\begin{figure}[tbh]
    \centering
    \includegraphics[width=0.9\linewidth]{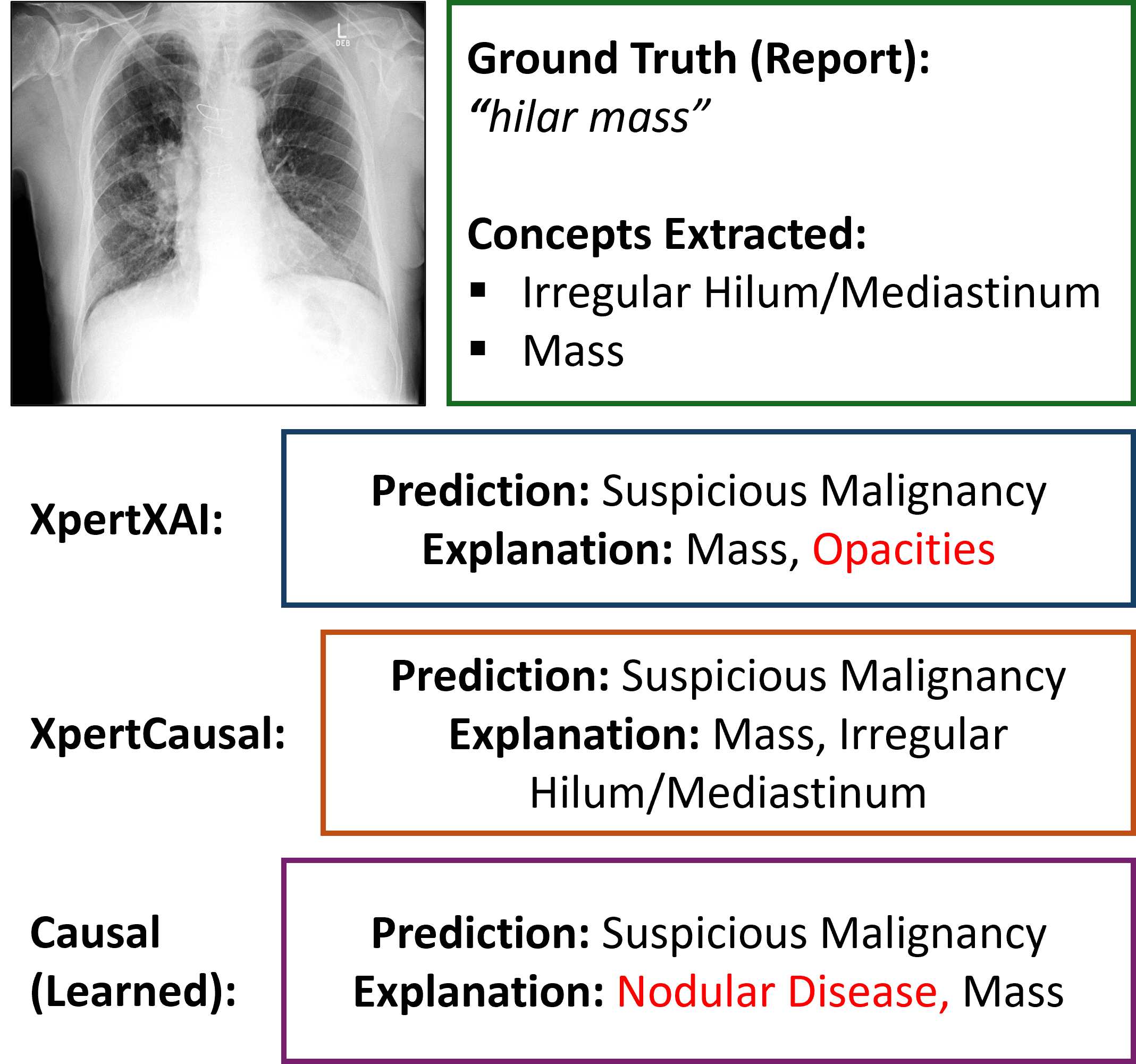}
    \caption{Example explanations from each approach for a CXR with a hilar mass. XpertXAI outputs the 2 concepts with the highest predicted probabilities. XpertCausal and the learned ablation model output the 2 concepts with the highest causal contribution. XpertCausal generates the explanation which is closest to the ground truth.}
    \label{fig:exps}
\end{figure}

All methods in this work follow a two-stage concept bottleneck pipeline, where a CXR is first mapped to a set of clinical concepts using a concept prediction model, and these concepts are then used to predict final pathology labels. To assess the impact of causal structure and expert knowledge, we compare three architectures: \textbf{XpertXAI}, a non-causal CBM; \textbf{XpertCausal}, our proposed causal model which incorporates radiologist-defined concept-pathology associations; and a causal ablation model (referred to as \textbf{Causal (Learned)}), where these associations are instead learned from data. An overview of each method is shown in Figure \ref{fig:overview}, with example explanations shown in Figure \ref{fig:exps}.

\subsection{Concept Prediction Model}

All models share the same concept prediction model, which maps input CXRs to a vector of clinical concepts. This model is based on an ImageNet-pretrained InceptionV3 architecture, taken from the original CBM work \cite{cbms}, and trained to predict the presence of each concept in a multi-label classification setting. Training is performed using a binary focal loss applied independently to each concept, with class-specific weighting to account for imbalance. Images are resized to $299 \times 299$ and normalised. The model is trained using stochastic gradient descent with momentum and weight decay, with a learning rate of $5 \times 10^{-4}$ and a batch size of 16. Training is run for up to 50 epochs, with early stopping based on validation performance. Downstream models use the predicted concept probabilities, while binary concept vectors are used only for evaluation. All experiments in this paper use an NVIDIA GTX 1060 6GB GPU.

\subsection{XpertXAI}

We use XpertXAI as our non-causal baseline. Specifically, we use the decision tree architecture, which was shown to perform best in the original work \cite{xpertxai}. This model maps predicted concepts to pathology labels using a learned set of decision rules, following a standard independent concept bottleneck framework \cite{cbms}. At inference time, explanations are derived directly from the concept prediction model. Concepts are ranked based on their predicted probabilities, and the top two concepts are reported as the explanation, following the original work.

\subsection{XpertCausal}
XpertCausal models the relationship between concepts and pathologies using a probabilistic causal architecture based on a radiologist-defined concept-pathology association matrix (Table \ref{tab:expert_matrix}), where each entry specifies whether a concept has a strong, weak, or no association with a given pathology. This matrix encodes these relationships in the pathology-to-concept direction, i.e. how the presence of a pathology gives rise to observable findings in an image. We note that this differs from typical approaches, which model the relationship in the reverse direction (Concept $\rightarrow$ Pathology), encoding concepts as features of pathologies rather than reflecting the underlying disease-to-finding generative process.

We model this generative process using a noisy-OR formulation, as multiple independent pathologies may each contribute to the presence of a radiographic finding. For each concept $c_k$, the probability of it being present given a set of pathologies $\mathbf{y}$ is defined as

\begin{align}
P(c_k = 1 \mid \mathbf{y}) = 1 - (1 - \lambda_k)\prod_{j : y_j = 1} (1 - \eta_{jk}),
\end{align}

where $\eta_{jk}$ represents the strength of association between pathology $j$ and concept $k$, and $\lambda_k$ is a leak term accounting for unexplained findings. The strength of each relationship is initialised from the expert-defined associations and then refined during training, while remaining constrained to only exist where defined by the expert matrix. Model parameters are refined by optimising binary cross-entropy loss.
At inference time, we invert this generative model to estimate pathology probabilities given the predicted concepts for a specific CXR. We use the InceptionV3 model's predicted concept probabilities during inference. Specifically, we compute

\begin{align}
P(\mathbf{y} \mid \mathbf{c}) = \frac{P(\mathbf{c} \mid \mathbf{y}) \, P(\mathbf{y})}{\sum_{\mathbf{y}'} P(\mathbf{c} \mid \mathbf{y}') \, P(\mathbf{y}')}
\end{align}

where $\mathbf{y}$ denotes a configuration of pathologies and $\mathbf{c}$ the predicted concept probabilities. The likelihood $P(\mathbf{c} \mid \mathbf{y})$ is defined by the noisy-OR model, and the prior $P(\mathbf{y})$ is estimated from the training data. As the number of pathologies is small, exact inference over all possible configurations is possible.

This framework reflects the structure of real-world diagnostic reasoning, where pathologies (e.g., pneumonia) cause observable findings on CXRs (e.g., opacities), from which final diagnoses are inferred. By explicitly modelling this direction and inverting it at inference time, XpertCausal provides a structured and clinically grounded probabilistic causal framework for modelling pathology-to-concept relationships and inferring pathologies from predicted concepts.
Explanations of model predictions are generated by measuring the contribution of each concept to the predicted pathologies using log-likelihood ratios, and selecting the top-$K$ supporting concepts.

\subsection{Causal (Learned)}

In order to properly isolate the effect of incorporating expert-defined relationships, we include a causal ablation model with the same probabilistic formulation as XpertCausal, but where concept-pathology associations are learned directly from data rather than specified by a radiologist. This model retains the same noisy-OR generative structure and Bayesian inference procedure, but does not constrain associations to those defined in the expert matrix. Instead, all concept-pathology relationships are treated as free parameters. These associations are initialised uniformly and are learned entirely during training, allowing the model to capture patterns present in the data without prior clinical constraints. Comparing this model to XpertCausal allows us to assess the impact of incorporating expert knowledge into the causal structure, while keeping the underlying modelling framework fixed. Explanations are generated in the same way as XpertCausal, by ranking concepts according to their contribution to the predicted pathologies using log-likelihood ratios.

\section{Results}

We evaluate the performance of XpertCausal through pathology prediction performance, concept prediction performance, explanation quality measured by top-$K$ concept overlap with ground truth, and the alignment of modelled concept-pathology associations with radiologist-defined relationships.

\subsection{Pathology Prediction Performance}\label{sec:label_perf}

The pathology prediction performance of XpertCausal and the Causal (Learned) ablation model is compared against XpertXAI and a baseline InceptionV3 classifier, providing comparison to both a non-causal concept-based model and a standard end-to-end baseline. Results are shown in Table \ref{tab:label_performance}, reported in terms of macro-averaged AUROC, F1 Score, and Expected Calibration Error (ECE). The InceptionV3 model serves as the image-to-label baseline, as all concept-based models in this work use an InceptionV3 architecture for their image-to-concept prediction stage.
XpertCausal achieves the highest performance across all three metrics, with an AUROC of 0.7968, F1 Score of 0.7109, and ECE of 0.0079. The Causal (Learned) ablation achieves lower performance, while XpertXAI achieves comparable F1 scores but lower AUROC and worse calibration. The InceptionV3 baseline performs lowest across all metrics. These results suggest that introducing causal structure into CBM design improves predictive performance and calibration, with further gains observed when directly incorporating expert-defined relationships.

\subsection{Concept Prediction Performance}\label{sec:concept_perf}

XpertXAI, XpertCausal, and the Causal (Learned) ablation all use the same InceptionV3-based concept prediction model, in order to ensure that differences in downstream performance arise from the label prediction architecture rather than the quality of concept extraction. Table \ref{tab:concept_performance} presents the performance of this shared concept encoder, reported in terms of both macro-averaged and per-concept AUROC and accuracy. The concept predictor achieves a macro-average AUROC of 0.8012 and accuracy of 0.8139 across all 11 concepts. Performance varies across concepts, with higher scores observed for more frequently occurring concepts such as \textit{Unremarkable}, and lower scores for rarer concepts such as \textit{Irregular Hilum/Mediastinum} and \textit{Consolidation}.

\begin{table}[tb]
    \centering
    \begin{tabular}{lrrr}
        \toprule
        \textbf{Model} & \textbf{AUROC} & \textbf{F1 Score} & \textbf{ECE} \\
        \midrule
        XpertCausal & \textbf{0.7968} & \textbf{0.7109} & \textbf{0.0079} \\
        Causal (Learned) & 0.7631 & 0.6846 & 0.0102 \\
        XpertXAI & 0.7361 & 0.6802 & 0.0249 \\
        InceptionV3 & 0.6833 & 0.6120 & 0.0377 \\
        \bottomrule
    \end{tabular}
    \caption{Macro-averaged pathology prediction performance for each model, reported as AUROC, F1 Score, and Expected Calibration Error (ECE). Lower ECE indicates better calibration.}
    \label{tab:label_performance}
\end{table}

\begin{table}[tb]
    \centering
    \begin{tabular}{lrr}
        \toprule
        \textbf{Concept} & \textbf{AUROC} & \textbf{Accuracy} \\
        \midrule
        Unremarkable & 0.9068 & 0.8627 \\
        Mass & 0.7978 & 0.7645 \\
        Nodular Disease & 0.8149 & 0.7872 \\
        Irregular Hilum/Mediastinum & 0.7300 & 0.9012 \\
        Pneumonitis & 0.7880 & 0.7823 \\
        Consolidation & 0.7246 & 0.8498 \\
        Opacities & 0.8095 & 0.7943 \\
        Infection & 0.7695 & 0.7109 \\
        Pleural Disease & 0.8418 & 0.8290 \\
        Enlarged Heart & 0.8653 & 0.8754 \\
        Absent Lung Markings & 0.7648 & 0.7958 \\
        \midrule 
        \textbf{Total} & 0.8012 & 0.8139 \\
        \bottomrule
    \end{tabular}
    \caption{Concept prediction performance of the shared InceptionV3-based concept model, reported as per-concept AUROC and accuracy. Total performance is macro-averaged across all concepts.}
    \label{tab:concept_performance}
\end{table}

\subsection{Explanation Quality}\label{sec:exp_quality}

\begin{figure}[tb]
    \centering
    \includegraphics[width=0.9\linewidth]{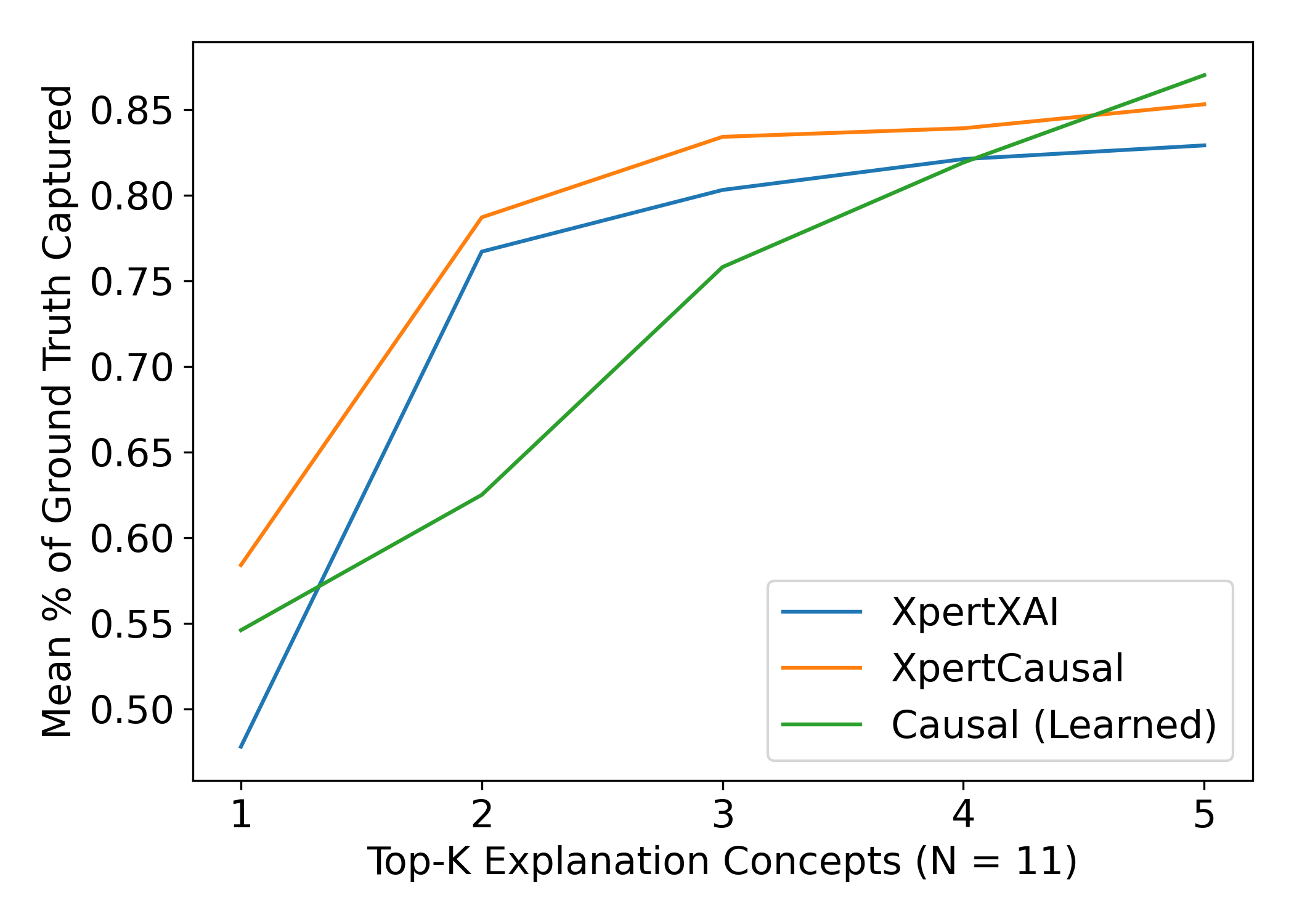}
    \caption{Mean proportion of ground truth concepts captured by the top-$K$ explanation concepts for each model, for $K=1$ to $5$. The full concept space contains 11 concepts.}
    \label{fig:explanation_topk}
\end{figure}

Figure \ref{fig:explanation_topk} shows the mean proportion of ground truth concepts captured by the top-$K$ explanation concepts for each model, for values of $K$ ranging from 1 to 5. Ground truth concepts are extracted from the radiology reports associated with each CXR, as described in Section \ref{sec:data}. For each model, explanation concepts are defined as the $K$ concepts with the highest contribution scores to the model’s predictions. For XpertXAI, concepts are ranked based on their predicted probabilities from the concept prediction model, while for XpertCausal and the Causal (Learned) ablation, concepts are ranked based on their contribution to the predicted labels, computed using the weighted concept-pathology associations encoded in their respective causal graphs. 

A concept is considered correctly identified if it appears in both the predicted explanation set and the ground truth concept set for a given CXR. As $K$ increases, a larger proportion of the fixed concept set ($N=11$) is included, and so the models begin to identify many of the same concepts. However, explanations are not intended to list all possible concepts. In practice, they should remain short and focused, reporting only the most diagnostically relevant findings. Therefore, strong performance at low $K$ is essential.

The largest differences between models are seen at lower values of $K$, while performance begins to converge as $K$ increases, with the Causal (Learned) ablation slightly exceeding the others at $K=5$. XpertCausal captures the highest proportion of ground truth concepts at low $K$, suggesting that it more reliably ranks diagnostically relevant concepts as the most important to model predictions. An example of this behaviour is shown in Figure \ref{fig:exps}.

\subsection{Graph Faithfulness}\label{sec:graph_faith}

To evaluate how well the learned concept-pathology relationships align with expert domain knowledge, we compare the association strengths of both causal models to the radiologist-defined matrix presented in Section \ref{sec:associations}. For each concept-pathology pair, edges are grouped according to their expert-defined category (Strong, Weak, or None), and the average learned edge weight is computed within each group.

Results are shown in Table \ref{tab:edge_weightings}. XpertCausal assigns the highest weights to edges labelled as Strong by the radiologist, with significantly lower weights for Weak associations and zero weights for edges labelled as None. In contrast, the Causal (Learned) ablation model produces less separation between categories, assigning higher average weights to Weak and None edges. This behaviour is further shown by specific high-weight edges learned by the Causal (Learned) model that are not present in the expert-defined matrix, including \textit{Pneumothorax} $\rightarrow$ \textit{Unremarkable} (0.4630), \textit{Suspicious Malignancy} $\rightarrow$ \textit{Unremarkable} (0.3374), and \textit{Cardiac Failure} $\rightarrow$ \textit{Unremarkable} (0.2843).

Overall, these results indicate that XpertCausal more closely reflects the expert-defined structure, while the unconstrained model learns additional associations which are not supported by real-world clinical reasoning.

\begin{table}[tb]
    \centering
    \begin{tabular}{lrrr}
        \toprule
        \textbf{Model} & \textbf{Strong} & \textbf{Weak} & \textbf{None} \\
        \midrule
        XpertCausal & 0.7268 & 0.0953 & 0.0000\\
        Causal (Learned) & 0.5783 & 0.3082 & 0.0920 \\
        \bottomrule
    \end{tabular}
    \caption{Average learned edge weights for concept-pathology associations, grouped by expert-defined categories (Strong, Weak, None). Values represent mean edge weight within each category.}
    \label{tab:edge_weightings}
\end{table}

\section{Discussion}

This work presents XpertCausal, a radiologist-guided causal CBM for CXR interpretation that combines concept bottleneck learning with probabilistic pathology-to-concept modelling. Unlike many structured or graph-based CBMs, XpertCausal models disease generation explicitly in the pathology-to-concept direction and performs Bayesian inversion over the generative model at inference time. Across pathology prediction, calibration, explanation quality, and alignment with expert-defined concept-pathology relationships, XpertCausal outperforms both the non-causal XpertXAI baseline and the unconstrained Causal (Learned) ablation model. These results suggest that incorporating clinically motivated causal structure into CBMs can improve not only predictive performance, but also the clinical plausibility of model reasoning.

The comparison between XpertCausal and the Causal (Learned) ablation highlights the importance of expert-defined structural constraints. Although both models use the same noisy-OR formulation and Bayesian inference procedure, the unconstrained model learns several high-weight associations that are clinically implausible, for example between pathology labels and the \textit{Unremarkable} concept. In contrast, XpertCausal preserves stronger separation between expert-defined Strong, Weak, and None associations. This suggests that causal structure alone is insufficient in noisy medical datasets: without constraints defined by human domain knowledge, learned graphs may reflect dataset-specific correlations rather than clinically meaningful disease-to-finding relationships.
In terms of the quality and clinical relevance of their explanations, XpertCausal achieves the highest top-$K$ overlap with ground truth concepts at low values of $K$, where explanations are most useful in practice. Since clinical explanations should be concise and focused, improvements at low $K$ are more important than performance at larger values of $K$, where much of the fixed concept set is eventually included. By ranking concepts according to their causal contribution to predicted pathologies, XpertCausal provides explanations which are more closely tied to the model's diagnostic inference process, rather than based on concept prediction confidence alone.

XpertCausal proposes an alternative framework for concept bottleneck modelling for medical imaging by reversing the conventional concept-to-pathology reasoning direction used in typical CBMs. Rather than treating concepts as independent predictive features used to infer disease (Concept $\rightarrow$ Pathology), XpertCausal instead models the clinically motivated generative process in which latent pathologies give rise to observable radiographic findings (Pathology $\rightarrow$ Concept). Pathology predictions are then obtained by performing Bayesian inference over this generative structure using the predicted concepts as evidence. This framework more closely reflects real clinical diagnostic reasoning, where diseases cause imaging findings that are interpreted by clinicians in order to reach diagnoses. By combining deep visual concept extraction with an explicit probabilistic pathology-to-finding model constrained by radiologist-defined associations, XpertCausal provides a structured and clinically grounded alternative to standard discriminative CBMs.

\subsubsection{Limitations and Future Work}

The radiologist-defined association matrix used in this work reflects a relatively limited concept space of 11 concepts and six target labels, and therefore may not capture the full range of clinically relevant findings and disease relationships present in real-world chest X-ray interpretation. We also note that both the concept definitions and the association matrix were designed by a single board-certified radiologist. However, as these concepts correspond to standard radiographic findings routinely used in clinical reporting, and individual radiologist reporting is standard clinical practice, the impact of this limitation is partially reduced. Future work will extend XpertCausal to larger concept and pathology spaces, incorporate additional expert-defined relationships, and evaluate generalisation across external datasets and imaging settings. An important next step is a formal radiologist evaluation of explanation quality for both XpertCausal and the unconstrained causal ablation model, assessing whether the generated explanations are clinically useful, trustworthy, and actionable within realistic diagnostic workflows.

\paragraph{Funding.} This work was supported by the Engineering and Physical Sciences Research Council (EPSRC) through the Causality in Healthcare AI Hub (CHAI), grant number EP/Y028856/1.

\appendix




\bibliographystyle{named}
\bibliography{ijcai26}

\end{document}